\documentclass{article}
\pdfoutput=1

\usepackage[final]{neurips_2021}
\usepackage
[acronym,smallcaps,nowarn,section,nogroupskip,nonumberlist]{glossaries}
\glsdisablehyper{}
\glsdisablehyper{}

\newacronym{KL}{KL}{Kullback-Leibler}
\newacronym{ELBO}{elbo}{\emph{evidence lower bound}}
\newacronym{MCMC}{mcmc}{Markov chain Monte Carlo}

\newacronym{ML}{ML}{machine learning}
\newacronym{VAE}{VAE}{variational auto-encoder}
\newacronym{VAETV}{VAE-TV}{variational auto-encoder with twin hidden variables}
\newacronym{AE}{AE}{auto-encoder}
\newacronym{SGD}{SGD}{stochastic gradient descent}
\newacronym[sort=beta]{BVAE}{\(\beta\)-vae}{}
\newacronym{TC}{TC}{total correlation}
\newacronym{DGM}{DGM}{deep generative model}
\newacronym{MMD}{MMD}{maximum mean discrepancy}
\newacronym{GAN}{GAN}{generative adversarial network}
\newacronym{AAE}{AAE}{adversarial auto-encoder}
\newacronym{CCM}{CCM}{constant curvature manifold}
\newacronym{IWAE}{IWAE}{importance weighted auto-encoder}
\newacronym{MC}{MC}{Monte Carlo}
\newacronym{NN}{NN}{neural network}
\newacronym{FNN}{FNN}{feedforward neural network}
\newacronym{MLP}{MLP}{multilayer perceptron}
\newacronym{SVM}{SVM}{support vector machine}
\newacronym{KDE}{KDE}{kernel density estimation}
\newacronym{RKHS}{RKHS}{reproducing kernel Hilbert space}
\newacronym{VI}{VI}{variational inference}
\newacronym{maxlike}{ML}{maximum likelihood}
\newacronym{NP}{NP}{neural process}
\newacronym{SP}{SP}{stochastic process}
\newacronym{ACNP}{ACNP}{attentive conditional neural process}
\newacronym{CNP}{CNP}{conditional neural process}
\newacronym{CNPs}{CNPs}{conditional neural processes}
\newacronym{SSL}{SSL}{self-supervised learning}
\newacronym{MSE}{MSE}{mean squared error}
\newacronym{CRESP}{CReSP}{Contrastive Representations of Stochastic Processes}
\newacronym{FCLR}{FCLR}{function contrastive learning}
\newcommand{\E}{\mathbb{E}}
\newcommand{\R}{\mathbb{R}}

\newcommand{\Z}{\mathcal{Z}}

\def\rvs{{\mathbf{s}}}

\usepackage{bm}
\newcommand{\C}{\mathcal{C}}
\newcommand{\z}{\bm{z}}
\newcommand{\bc}{\bm{c}}
\newcommand{\br}{\bm{r}}
\newcommand{\bnew}{\bm{c}^\star}
\newcommand{\znew}{\z^\star}
\newcommand{\bu}{\bm{u}}
\newcommand{\bv}{\bm{v}}

\newcommand{\cov}{\bm{x}}
\newcommand{\covnew}{\cov^\star}
\newcommand{\Cov}{\mathcal{X}}
\newcommand{\obs}{\bm{y}}
\newcommand{\obsnew}{\obs^\star}
\newcommand{\Obs}{\mathcal{Y}}
\newcommand{\lab}{\ell}
\newcommand{\Lab}{L}

\usepackage[utf8]{inputenc} %
\usepackage[T1]{fontenc}    %
\usepackage{hyperref}       %
\usepackage[capitalise,noabbrev]{cleveref}
\usepackage{url}            %
\usepackage{booktabs}       %
\usepackage{amsfonts}       %
\usepackage{nicefrac}       %
\usepackage{microtype}      %

\usepackage{nccmath}

\usepackage[disable]{todonotes}
\usepackage{menukeys}
\usepackage{multirow}
\usepackage{graphicx}
\usepackage{mathrsfs}
\usepackage{paralist}
\usepackage{subcaption}
\usepackage{amsthm}
\usepackage{enumitem}
\usepackage{xspace}

\usepackage{amsmath,amssymb}
\usepackage[ruled]{algorithm2e}

\Crefname{equation}{Eq.}{Eqs.}
\Crefname{figure}{Fig.}{Figs.}
\Crefname{table}{Tab.}{Tabs.}
\Crefname{section}{Sec.}{Secs.}

\usepackage{xcolor}

\newcommand\Todo[1]{\textcolor{red}{\\Todo: #1}}

\usepackage{float}

\usepackage[
  separate-uncertainty = true,
  multi-part-units = repeat
]{siunitx}

\usepackage{wrapfig}
\usepackage{pdfcomment}
\usepackage{booktabs, tabularx}
\hypersetup{
  colorlinks,
  linkcolor={red!50!black},
  citecolor={blue!50!black},
  urlcolor={blue!80!black}
}
\definecolor{orange}{HTML}{FF7F0E}
\definecolor{blue}{HTML}{1F77B4}
\definecolor{green}{HTML}{2ca02c}

\newcommand*\samethanks[1][\value{footnote}]{\footnotemark[#1]}

\usepackage[title]{appendix}

\title{On Contrastive Representations \\ of Stochastic Processes}

\author{
  Emile Mathieu$^{\dagger}$\thanks{Equal contribution. Author ordering determined by coin flip.}\ , \ 
  Adam Foster$^{\dagger}$\samethanks\ , \ 
  Yee Whye Teh$^{\dagger, \ddagger}$ \\
  \texttt{\{emile.mathieu, adam.foster, y.w.teh\}@stats.ox.ac.uk}, \\
  $\dagger$ Department of Statistics, University of Oxford, United Kingdom\\
  $\ddagger$ DeepMind, United Kingdom\\
}

\hypersetup{
  pdfauthor={Emile Mathieu, Adam Foster, Yee Whye Teh},
  pdftitle={On Contrastive Representations of Stochastic Processes}
}

\begin{document}

\maketitle
\renewcommand{\thefootnote}{\arabic{footnote}}
\setcounter{footnote}{0}

\begin{abstract}

Learning representations of stochastic processes is an emerging problem in machine learning with applications from meta-learning to physical object models to time series. Typical methods rely on exact reconstruction of observations, but this approach breaks down as observations become high-dimensional or noise distributions become complex. To address this, we propose a unifying framework for learning contrastive representations of stochastic processes (\acrshort{CRESP}) that does away with exact reconstruction. We dissect potential use cases for stochastic process representations, and propose methods that accommodate each. Empirically, we show that our methods are effective for learning representations of periodic functions, 3D objects and dynamical processes. Our methods tolerate noisy high-dimensional observations better than traditional approaches, and the learned representations transfer to a range of downstream tasks.
\end{abstract}

\renewcommand\tabularxcolumn[1]{m{#1}}
\newcolumntype{C}{>{\centering\arraybackslash}X}
\begin{wraptable}{r}{0.45\columnwidth}
	\vspace{-1.3em}
	\caption{Example stochastic processes with covariate space $\Cov$ and observation space $\Obs$.}
	\label{tab:stochatic_process}
	\small
	\renewcommand{\arraystretch}{1.}
	\begin{tabularx}{0.45\columnwidth}{>{\hsize=0.02\hsize}C
			>{\hsize=0.18\hsize}C
			>{\hsize=0.18\hsize}C
			>{\hsize=0.62\hsize}C}                        
		 & $\Cov$ & $\Obs$ & Illustration \\
		\midrule
		\rotatebox[origin=c]{90}{\tiny 1D function} & $\R$  & $\R$    & \includegraphics[width=\hsize]{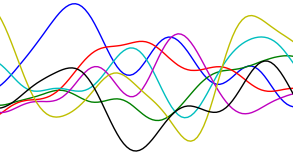}                       \\
		\rotatebox[origin=c]{90}{\tiny Image in-fill} & $\mathbb{Z}^2$  & $\R^3$    & \includegraphics[width=\hsize]{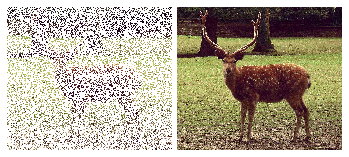}                       \\
		\rotatebox[origin=c]{90}{\tiny 3D object} & $SE(3)$ & Images & \includegraphics[width=\hsize]{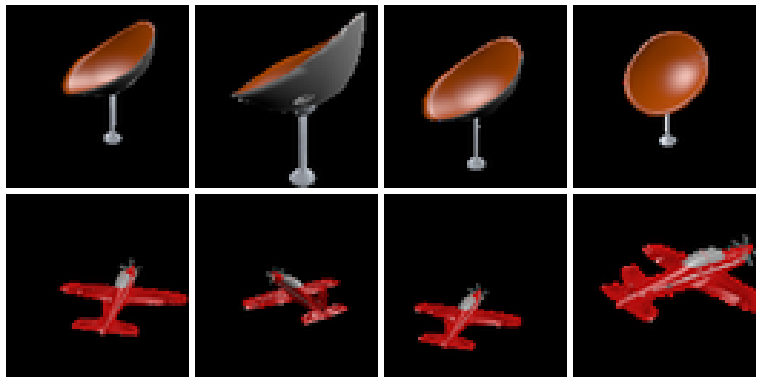} \\
		\rotatebox[origin=c]{90}{\tiny Video} & $\R$  & Images   & \includegraphics[width=\hsize]{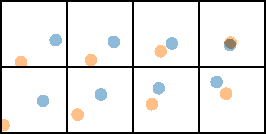}
	\end{tabularx}
\end{wraptable}
\section{Introduction}
\label{sec:intro}
The stochastic process \citep{doob1953stochastic,parzen1999stochastic} is a powerful mathematical abstraction used in biology \citep{bressloff2014stochastic}, chemistry \citep{van1992stochastic}, physics \citep{jacobs2010stochastic}, finance \citep{steele2012stochastic} and other fields.
The simplest incarnation of a stochastic process is a random function $\R\to\R$, such as a Gaussian Process \citep{mackay2003information}, that can be used to describe a real-valued signal indexed by time or space. 
Extending to random functions from $\R$ to another space, stochastic processes can model time-dependent phenomena like queuing \citep{grimmett2020probability} and diffusion \citep{ito2012diffusion}.
In meta-learning, the stochastic process can be used to describe few-shot learning tasks---mappings from images to class labels \citep{vinyals2016matching}---and image completion tasks---mappings from pixel locations to RGB values \citep{garnelo2018conditional}.
In computer vision, 2D views of 3D objects can be seen as observations of a stochastic process indexed by the space of possible viewpoints \citep{eslami2018Neural,mildenhall2020nerf}. Videos can be seen as samples from a time-indexed stochastic process with 2D image observations \citep{zelnik2001event}.

\todo{yw: stretching it somewhat}
Machine learning algorithms that operate on data generated from stochastic processes are therefore in high demand.
We assume that we have access to only a small set of covariate--observation pairs $\{(\cov_i,\obs_i)_{i=1}^C\}$ from different realizations of the underlying stochastic process.
This might correspond to a few views of a 3D object, or a few snapshots of a dynamical system evolving in time.
Whilst conventional deep learning thrives when there is a large quantity of i.i.d.~data available \citep{lake2017building}, allowing us to learn a fresh model for each realization of the stochastic process, when the context size is small it makes sense to use data from other realizations to build up prior knowledge about the domain which can aid learning on new realizations \citep{reed2017few,garnelo2018conditional}.

Traditional methods for learning from stochastic processes, including the Gaussian Process family \citep{mackay2003information,rasmussen2003gaussian} and the Neural Process family \citep{garnelo2018conditional,garnelo2018neural,eslami2018Neural}, learn to reconstruct a realization of the process from a given context.
That is, given a context set $\{(\cov_i,\obs_i)_{i=1}^C\}$, these methods provide a predictive distribution $q\left(\obsnew|\covnew, (\cov_i,\obs_i)_{i=1}^C\right)$ for the observation that would be obtained from this realization of the process at any target covariate $\covnew$. These methods use an explicit likelihood for $q$, typically a Gaussian distribution. Whilst this can work well when $\obsnew$ is low-dimensional and unimodal, it is a restrictive assumption. For example, when $p\left(\obsnew|\covnew, (\cov_i,\obs_i)_{i=1}^C\right)$ samples a high-dimensional image with colour distortion, traditional methods must learn to perform conditional image generation, a notably challenging task \citep{oord2016conditional,chrysos2021cope}.

In this paper, we do away with the explicit likelihood requirement for learning from stochastic processes.
Our first insight is that, for a range of important downstream tasks, exact reconstruction is not necessary to obtain good performance.
Indeed, whilst $\obs$ may be high-dimensional, the downstream target label or feature $\lab$ may be simpler.
We consider two distinct settings for $\lab \in \Lab$. The first is a downstream task that \emph{depends on the covariate $\cov \in \Cov$}, formally a second process $\Cov\to L$ that covaries with the first. For example, $\lab(\cov)$ could represent a class label or annotation for each video frame.
The second is a downstream task that depends on the entire process realization, such as a single label for a 3D object.
In both cases, we assume that we have limited labelled data, so we are in a semi-supervised setting \citep{zhu2005semi}.

To solve problems of this nature, we propose a general framework for \gls{CRESP}.
At its core, \gls{CRESP} consists of a flexible encoder network architecture for contexts $\{(\cov_i,\obs_i)_{1}^C\}$ that unites transformer encoders of sets \citep{vaswani2017attention,parmar2018image} with convolutional encoders \citep{lecun1989backpropagation} for observations that are images.
To account for the two kinds of downstream task that may of interest, we propose a \emph{targeted} variant of \gls{CRESP} that learns a representations depending on the context and a target covariate $\covnew$, and an \emph{untargeted} variant that learns one representation of the context.
To train our encoder, we take our inspiration from recent advances in contrastive learning \citep{bachman2019learning,chen2020simple} which have so far focused on representations of single observations, typically images.
We define a variant of the InfoNCE objective \citep{oord2018representation} for contexts sampled from stochastic processes, allowing us to avoid training objectives that necessitate exact reconstruction.
Rather than attempting pixel-perfect reconstruction, then, \gls{CRESP} solves a self-supervised task in representation space.

The \gls{CRESP} framework unifies and extends recent work, building on \gls{FCLR} \citep{gondal2021Function} by considering targeted as well as untargeted representations and using self-attention in place of mean-pool aggregation.
We develop on noise contrastive meta-learning \citep{ton2019noise} by focusing on downstream tasks rather than multi-modal reconstruction, replacing conditional mean embeddings with neural representations and using a simpler training objective.

We evaluate \gls{CRESP} on sinusoidal functions, 3D objects, and dynamical processes with high-dimensional observations.
We empirically show that our methods can handle high-dimensional observations with naturalistic distortion, unlike explicit likelihood methods, and our representations lead to improved data efficiency compared to supervised learning.
\gls{CRESP} performs well on a range of downstream tasks, both targeted and untargeted, outperforming existing methods across the board.
Our code is publicly available at \url{github.com/ae-foster/cresp}.

\section{Background}
\label{sec:background}

\paragraph{Stochastic Processes}
\label{sec:sps}
Stochastic Processes (SPs) are probabilistic objects defined as a family of random variables indexed by a covariate space $\Cov$. 
For each $\cov\in\Cov$, there is a corresponding random variable $\obs|\cov \in\Obs$ living in the observation space.
For example, $\cov$ might represent a pose and $\obs$ a photograph of an underlying object take from pose $\cov$ (see~\cref{tab:stochatic_process}).
We assume that there is a \emph{realization} $F$ sampled from a prior $p(F)$, and that the random variable $\obs|\cov$ is a sample from $p(\obs|F,\cov)$.
Thus, for each $\cov\in\Cov$, $F$ defines a conditional distribution $p(\obs|F,\cov)$.
We assume that observations are independent conditional on the realization $F$.
Hence, the joint distribution of multiple observations at locations $\cov_{1:C}$ from one realization of the stochastic process with prior $p(F)$ is
\begin{equation} \label{eq:stochastic_process}
    p(\obs_{1:C}|\cov_{1:C}) = \int p(F)\prod_{i=1}^C p(\obs_i|F, \cov_i) \ dF.
\end{equation}
Conversely, assuming \emph{exchangeability} and \emph{consistency}, the Kolmogorov Extension Theorem guarantees that the joint distribution takes the form \eqref{eq:stochastic_process} \citep{oksendal2003stochastic,garnelo2018neural}.

\paragraph{Neural Processes}

The neural process (NP) and conditional neural process (CNP) are closely related models that learn representations of data generated by a \acrfull{SP}\footnote{Note that neither the \gls{NP} nor the \gls{CNP} is formally \glspl{SP} as they do \emph{not} satisfy the consistency property.}.
The training objective for the NP and CNP is inspired by the posterior predictive distribution for SPs: given a context $\{(\cov_i,\obs_i)_{i=1}^C\}$, the observation at the target covariate $\covnew$ has the distribution
\begin{equation}
\label{eq:true_posterior_predictive}
\begin{split}
p(\obsnew|\covnew, (\cov_i,\obs_i)_{i=1}^C) = \int p(F|(\cov_i,\obs_i)_{i=1}^C) ~p(\obsnew|F, \covnew) \ dF.
\end{split}
\end{equation}
The CNP learns a neural approximation $q\left(\obsnew|\covnew, (\cov_i,\obs_i)_{i=1}^C\right) = p(\obsnew|\bc,\covnew)$ to equation \eqref{eq:true_posterior_predictive}, where $\bc = \sum_{i} g_\text{enc}(\cov_i,\obs_i)$ is a permutation-invariant \emph{context representation} and $p(\cdot|\bc,\cov)$ is an explicit likelihood.
Conventionally, $p$ is a Gaussian with mean and variance given by a neural network applied to $\bc,\cov$.
The CNP model is then trained by maximum likelihood.
In the NP model, an additional latent variable $\bu$ is used to represent uncertainty in the process realization, more closely mimicking \eqref{eq:true_posterior_predictive}.

A significant limitation, common to the NP family, is the reliance on an explicit likelihood.
Indeed, requiring $\log q\left(\obsnew|\covnew, (\cov_i,\obs_i)_{i=1}^C\right)$ to be large requires the model to successfully reconstruct $\obsnew$ based on the context, similarly to the reconstruction term in variational autoencoders \citep{kingma2014auto}.
Furthermore, the NP objective cannot be increased by extracting additional features from the context unless the predictive part of the model, the part mapping from $(\bc,\cov)$ to a mean and variance, is powerful enough to use them.

\paragraph{Contrastive Learning and Likelihood-free Inference}
Contrastive learning has enjoyed recent success in learning representations of high-dimensional data \citep{oord2018representation,bachman2019learning,he2020momentum,chen2020simple}, and is deeply connected to likelihood-free inference \citep{gutmann2010noise,oord2018representation,durkan2020contrastive}.
In its simplest form, suppose we have a distribution $p(\obs,\obs')$, for example $\obs$ and $\obs'$ could be differently augmented versions of the same image.
Rather than fitting a model to predict $\obs'$ given $\obs$, which would necessitate high-dimensional reconstruction, contrastive learning methods can be seen as learning the likelihood-ratio
$r(\obs'|\obs) = p(\obs'|\obs) / p(\obs')$.
To achieve this, contrastive methods encode $\obs,\obs'$ to deterministic embeddings $\z,\z'$, and consider additional `negative' samples $\z'_1,...,\z'_{K-1}$ which are the embeddings of other independent samples of $p(\obs')$ (for example, taken from the same training batch as $\obs,\obs'$).
The InfoNCE training loss \citep{oord2018representation} is then given by
\begin{equation} \label{eq:contrastive}
\mathcal{L}^{\text{InfoNCE}}_K = -\E\left[\log \frac{s(\z, \z')}{s(\z,\z') + \sum_k s(\z, \z'_k)}\right] - \log K.
\end{equation}
for similarity score $s>0$.
Informally, InfoNCE is minimized when $\z$ is more similar to $\z'$---the `positive' sample---than it is to the negative samples $\z'_1,...,\z'_{K-1}$ that are independent of $\z$.
Formally, \cref{eq:contrastive} is the multi-class cross-entropy loss arising from classifying the positive sample correctly.
It can be shown that the optimal similarity score $s$ is proportional to the true likelihood ratio $r$ \citep{oord2018representation,durkan2020contrastive}.
A key feature of InfoNCE is that learns about the predictive density $p(\obs'|\obs)$ indirectly, rather than by attempting direct reconstruction.

\section{Method} \label{sec:method}

Given data $\{(\cov_i,\obs_i)_{i=1}^C\}$ sampled from a realization of a stochastic process, one potential task is to make predictions about how observations will look at another $\covnew$---this is the task that is solved by the NP family.
However, in practice the inference that we want to make from the context data could be different.
For instance, rather than predicting a high-dimensional observation at a future time or another location, we could be interested in inferring some low-dimensional feature of that observation---whether two objects have collided at that point in time, 
or if an object can be seen from a given pose.
Even more simply, we might be solely interested in classifying the context, deciding what object is being viewed, for example.
Such \emph{downstream tasks} provide a justification for learning representations of stochastic processes that are not designed to facilitate predictive reconstruction of the process at some $\covnew$.
We break downstream tasks for stochastic processes into two categories.

\paragraph{Targeted and untargeted tasks}
A \emph{targeted} task is one in which the label $\lab$ depends on $\cov$, as well as on the underlying realization of the process $F$.
This means that we augment the stochastic process of \cref{sec:sps} by introducing a conditional distribution $p(\lab|F,\cov)$.
The goal is to infer the predictive density $p\left(\lab^\star|\covnew, (\cov_i,\obs_i)_{i=1}^C\right)$.
An \emph{untargeted task} associates one label $y$ with the entire realization $F$ via a conditional distribution $p(\lab|F)$.
The aim is to infer the conditional distribution $p\left(\lab| (\cov_i,\obs_i)_{i=1}^C\right)$.

\paragraph{Representation learning}
We assume a semi-supervised \citep{zhu2005semi} setting, with unlabelled contexts for a large number of realizations of the stochastic process, but few labelled realizations.
To make best use of this unlabelled data, we learn representations of contexts, and then fit a downstream model on top of fixed representations.
In the stochastic process context, we have the requirement for a representation learning approach that can transfer to both targeted and untargeted downstream tasks.
We therefore propose a general framework to learn contrastive representations of stochastic processes (\gls{CRESP}).
Our framework consists of a flexible encoder architecture that processes the context $\{(\cov_i,\obs_i)_{i=1}^C \}$ and a $\covnew$-dependent head for targeted tasks. This means \gls{CRESP} can encode data from stochastic processes in two ways: 1) a \emph{targeted} representation that depends on the context $\{(\cov_i,\obs_i)_{i=1}^C \}$ and some target location $\covnew$, being a predictive representation for the process at this covariate, suitable for targeted downstream tasks; or 2) a single \emph{untargeted} representation of the context $\{(\cov_i,\obs_i)_{i=1}^C \}$ that summarizes the entire realization $F$, suitable for untargeted tasks.

\subsection{Training} \label{sec:training}
We have unlabelled data $\{(\cov_i,\obs_i)_{i=1}^C \}$ that is generated from the stochastic process \eqref{eq:stochastic_process}, but unlike the Neural Process family, we do not wish to place an explicit likelihood on the observation space $\Obs$.
Instead, we adopt a contrastive self-supervised learning approach \citep{oord2018representation,bachman2019learning,chen2020simple} to training.
Whilst we adopt subtly different training schemes for the targeted and untargeted cases, the broad strokes are the same.
Given a mini-batch of contexts samples from different realizations of the underlying stochastic process, create predictive and ground truth representations from each.
We then use representations from other observations in the same mini-batch as negative samples in an InfoNCE-style~\citep{oord2018representation} training loss.
This can be seen as learning an \emph{unnormalized} likelihood ratio.
Taking gradients through this loss function allows us to update our \gls{CRESP} network by gradient descent \citep{robbins1951stochastic}.
We now describe the key differences between the \emph{targeted} and \emph{untargeted} cases. %

\begin{figure}[t]
	\centering
	\begin{subfigure}[b]{0.49\linewidth}
		\centering
		\includegraphics[trim=0 0 290 0,clip,width=\textwidth]{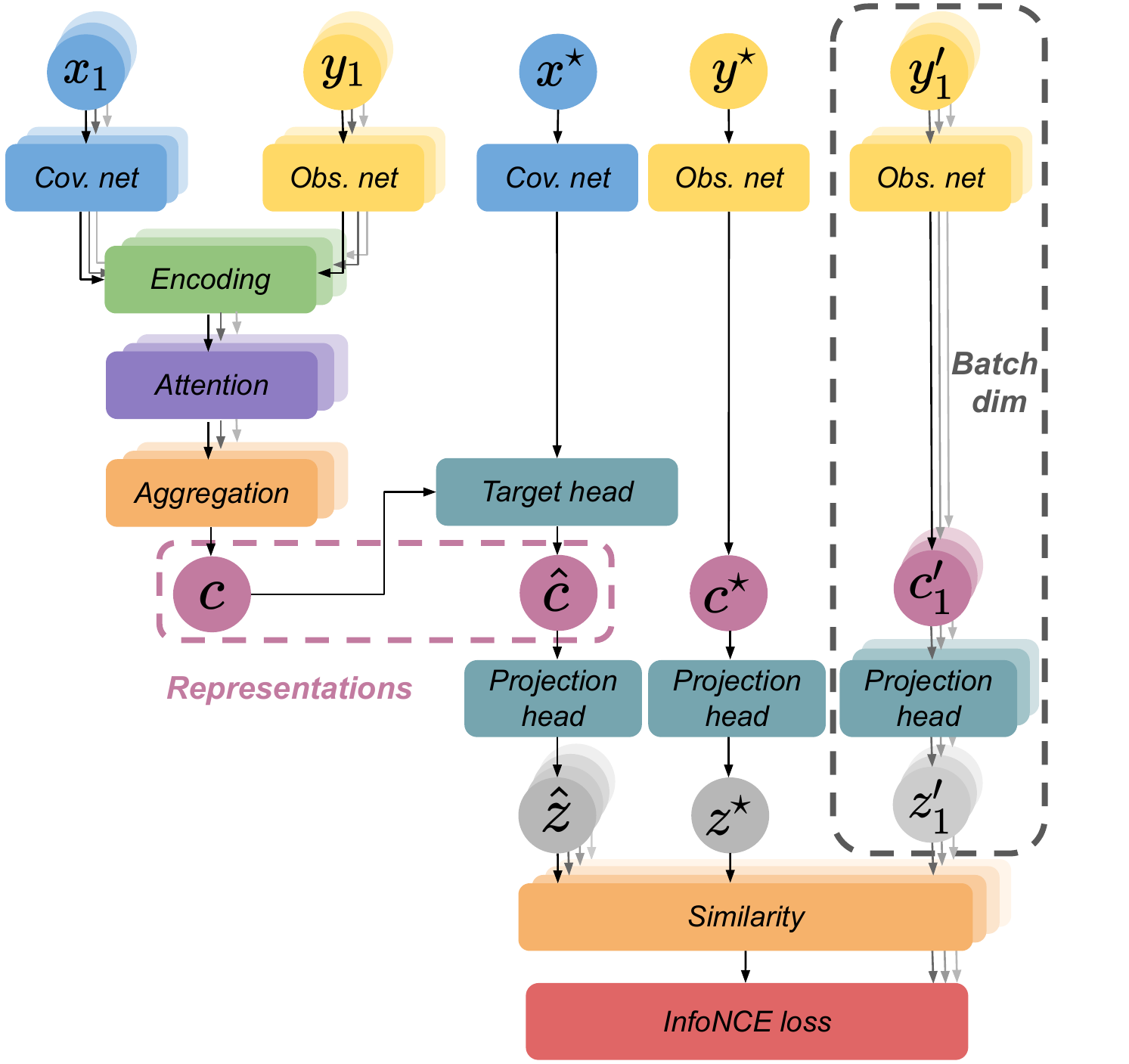}
		\label{fig:targeted_diag}
	\end{subfigure}
	\hfill
	\begin{subfigure}[b]{0.50\linewidth}
		\centering
		\includegraphics[trim=0 0 250 0,clip,width=\textwidth]{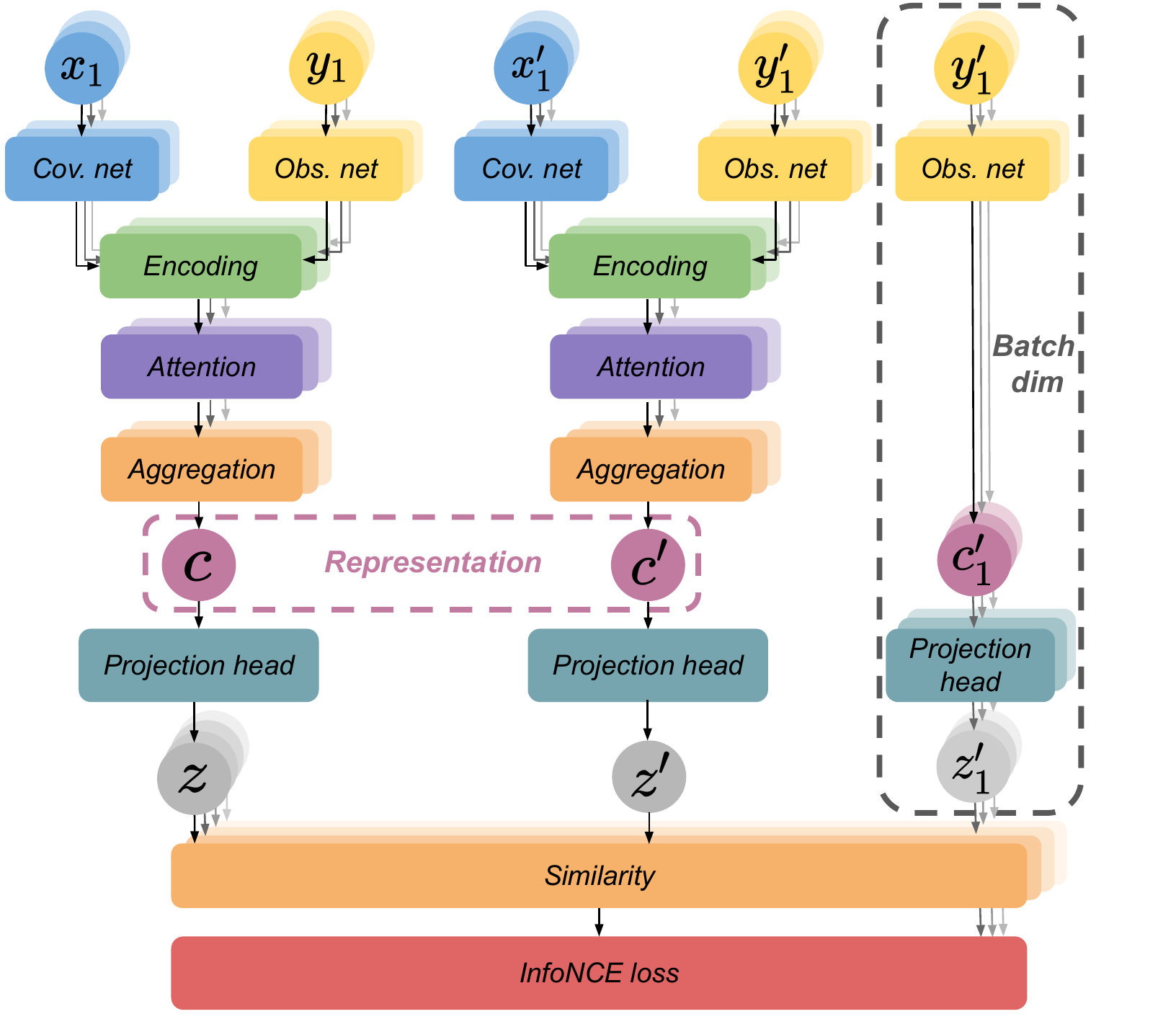}
		\vspace{1.em}
		\label{fig:untargeted_diag}
	\end{subfigure}
	\caption{\gls{CRESP} architecture with contrastive loss. [Left] Targeted, [Right] Untargeted.}
	\label{fig:graphical_model}
\end{figure}

\paragraph{Targeted CReSP}
This setting is closer in spirit to the CNP.
Rather than making a direct estimate of the posterior predictive $p(\obsnew|\covnew,(\cov_i,\obs_i)_{i=1}^C)$ for each value of $\covnew$, we instead attempt to learn the following likelihood-ratio 
\begin{equation}
	\label{eq:targeted_lr}
	r(\obsnew|\covnew, (\cov_i,\obs_i)_{i=1}^C) = \frac{p(\obsnew|\covnew,(\cov_i,\obs_i)_{i=1}^C)}{p(\obsnew)}
\end{equation}
where $p(\obsnew)$ is the marginal distribution of observations from different realizations of the process and different covariates.
To estimate this ratio with contrastive learning, we first randomly separate the context $\{(\cov_i,\obs_i)_{i=1}^C \}$ into a training context $\{(\cov_i,\obs_i)_{i=1}^{C-1} \}$ and a target $(\covnew,\obsnew)$.
We then process $\{(\cov_i,\obs_i)_{i=1}^{C-1}, \covnew\}$ and $\obsnew$ separately with an encoder network, yielding respectively a predictive representation $\hat{\bc}$ and a target representation $\bnew$.
This encoder network is described in detail in the following \cref{sec:representation}.
These representations are further projected into a low-dimensional space $\Z$ using a shallow MLP, referred as \texttt{Projection head} on \cref{fig:graphical_model}, giving $\hat{\z}$ and $\znew$.
We create negative samples $\z_1',...,\z_{K-1}'$, defined as samples coming from other realisations of the stochastic process, from representations obtained from the other observations of the batch.
This means that we are drawing negative samples via the distribution $p(\obsnew)$ as required for \eqref{eq:targeted_lr}.
We then form the contrastive loss
\begin{equation} \label{eq:targeted_cresp_loss}
	\mathcal{L}^\text{targeted}_K = -\E \left[ \log \frac{s(\znew, \hat{\z})}{s(\znew, \hat{\z}) + \sum_k s(\z_k', \hat{\z})} \right]- \log K
\end{equation}
with $s(\znew, \hat{\z}) =\exp \left(\z^{\star\top} \hat{\z}/\tau\|\z^\star\|\|\hat{\z}\| \right)$.
By minimizing this loss, we ensure that the predicted representation is closer to the representation of the true outcome than representations of other random outcomes. The optimal value of $s(\znew, \hat{\z})$ is proportional to the likelihood ratio $r(\obsnew|\covnew,(\cov_i,\obs_i)_{i=1}^C)$.
\paragraph{Untargeted CReSP}
For the untargeted version, we simply require a representation of each context; $\covnew$ no longer plays a role. The key idea here is that, without estimating a likelihood ratio in $\Obs$ space, we can use contrastive methods to encourage two representations formed from the \emph{same realization} of the stochastic process to be more similar than representations formed from 
\emph{different} realizations. %
To achieve this, we randomly split the whole context $\{(\cov_i,\obs_i)_{i=1}^C \}$ into two training contexts $\{(\cov_i,\obs_i)_{i=1}^{C_1}\}$ and $\{(\cov_i',\obs_i')_{i=1}^{C_2}\}$, with an equal split $C_1=C_2=C/2$ being our standard approach.
We encode both with an encoder network, giving two representations $\bc,\bc'$, further projected into lower-dimensional representations $\z,\z'$ as in the targeted case.
We also take $K-1$ negative samples $\z'_1,...,\z'_K$ using other representations in the same training mini-batch. 
\begin{equation} \label{eq:untargeted_cresp_loss}
	\mathcal{L}_K^\text{untargeted} = -\E\left[ \log\frac{s(\z, \z')}{s(\z,\z') +  \sum_k s(\z,\z'_k)} \right] - \log K.
\end{equation}
This training method is closer in spirit to SimCLR~\citep{chen2020simple}, but here we include attention and aggregation steps to combine the distinct elements of the context.

\subsection{Representation} \label{sec:representation}
The core of our architecture is a flexible encoder of a context $\{(\cov_i,\obs_i)_{i=1}^C \}$, as illustrated in \cref{fig:graphical_model}.
	\paragraph{Covariate and observation preprocessing}
	We begin by applying separate networks to the covariate $g_\text{cov}(\cov)$ and observation $g_\text{obs}(\obs)$ of each pair $(\cov, \obs)$ of the context. 
	When observations $\obs$ are high-dimensional, such as images, this step is crucial because we can use existing well-developed vision architectures such as CNNs \citep{lecun1989backpropagation} and ResNets \citep{he2016deep} to extract image features.
	For covariates that are angles, we use Random Fourier Features \citep{rahimi2007random}.
	\paragraph{Pair encoding}
	We then combine separate encodings of $\cov,\obs$ into a single representation for the pair.
	We concatenate the individual representations and pass them through a simple neural network, i.e.\ $g_\text{enc}(\cov,\obs) := g_\text{enc}([g_\text{cov}(\cov), g_\text{obs}(\obs)])$.
	In practice, we found that a gated architecture works well.
	\paragraph{Attention \& Aggregation}
	We apply self-attention \citep{vaswani2017attention} over the $C$ different encodings of the context
	$\{g_\text{enc}(\cov_i,\obs_i)\}_{i=1}^C$.
	We found transformer attention \citep{parmar2018image} to perform best.
	We then pool the $C$ reweighted representations
	to yield a single representation $\bc = \sum_{i} g_\text{enc}(\cov_i,\obs_i)$.
	For targeted representations, we concatenate $\bc$ and $\covnew$, then pass them through a \emph{ target head}
	yielding $\hat{\bc} = h(\cov^\star, \bc)$, the predictive representation at $\covnew$.

\subsection{Transfer to downstream tasks} \label{sec:transfer}
We have outlined the \emph{unsupervised} part of \gls{CRESP}---a way to learn a representation of a context sampled from a stochastic process without explicit reconstruction. We now return to our core motivation for such representations, which is to use them to solve a downstream task, either targeted or untargeted.
This will be particularly useful in a \emph{semi-supervised} setting, in which labelled data for the downstream task is limited compared to the unlabelled data used for unsupervised training of the \gls{CRESP} encoder.
Our general approach to both targeted and untargeted downstream tasks is to fit \emph{linear} models on the context representations of the labelled training set, and use these to predict labels on new, unseen realizations of the stochastic process, following the precedent in contrastive learning \citep{hjelm2018learning,kolesnikov2019revisiting}. We do not use fine-tuning.

For targeted tasks, we assume that we have labelled data from $n$ realizations of the stochastic process that takes the form of an \emph{unlabelled} context $(\cov_{ij},\obs_{ij})_{i=1}^{C}$ along with a \emph{labelled pair} $(\cov_j^\star, \lab_j^\star)$ for each $j=1,\dots,n$. Here, $\lab_j^\star$ is the label at location $\cov^\star$ for realization $j$.
To fit a downstream classifier using \gls{CRESP} representations with this labelled dataset, we first process each $(\cov_{ij},\obs_{ij})_{i=1}^{C}$ along with the covariate $\cov_j^\star$ through a \emph{targeted} \gls{CRESP} encoder to produce $\hat{\bc}_j$.
This allows us to form a training dataset $(\hat{\bc}_j,\lab^\star_j)_{j=1}^{n}$ of representation, label pairs which we then use to train our downstream classifier.
At test time, given a test context $(\cov_{i}',\obs_{i}')_{i=1}^{C}$, we can predict the unknown label at any $\cov^\star$ by forming the corresponding targeted representation with the \gls{CRESP} network, and then feeding this into the linear classifier.
This is akin to zero-shot learning \citep{xian2018zero}.

For untargeted tasks, the downstream model is simpler. Given labelled data consisting of contexts $(\cov_{ij},\obs_{ij})_{i=1}^{C}$ with label $\lab_j$ for $j=1,\dots,n$, we can use the untargeted \gls{CRESP} encoder to produce a training dataset $(\bc_j,\lab_j)$ as before.
Actually, \emph{targeted} \gls{CRESP} can also be used to obtain \emph{untargeted} representations $\bc_j$--- without applying the target head.
We then use this to train the linear classifier. At test time, we predict labels for contexts from new, unseen realizations of the stochastic process.

\section{Related work}
\label{sec:relatedwork}

\paragraph{Neural process family}
Neural Processes \citep{garnelo2018neural} and Conditional Neural Processes \citep{eslami2018Neural,garnelo2018conditional} are closely related methods that create a representation of an stochastic process realization by aggregating representations of a context.
Unlike \gls{CRESP}, NPs are generative models that uses an explicit likelihood, generally a fully factorized Gaussian, to estimate the posterior predictive distribution.
Attentive (Conditional) Neural Processes \citep[A(C)NP]{kim2019attentive} introduced both self-attention and cross-attention into the NP family. 
The primary distinction between this family and \gls{CRESP} is the explicit likelihood that is used for reconstruction.
As the most comparable method to \gls{CRESP}, we focus on the (A)CNP in the experiments.

\paragraph{SimCLR family}
Recent popular methods in contrastive learning \citep{oord2018representation,bachman2019learning,tian2020Contrastive,chen2020simple} create neural representations of single objects, typically images, that are approximately invariant to a range of transformations such as random colour distortion.
Like \gls{CRESP}, many of these approaches use the InfoNCE objective to train encoders.
What distinguishes \gls{CRESP} from conventional contrastive learning methods is that it provides representations of realizations of stochastic processes, rather than of individual images. Thus, standard contrastive learning solves a strictly less general problem than \gls{CRESP} in which the covariate $\cov$ is absent. Standard contrastive encoders do not aggregate multiple covariate-observation pairs of a context, although simpler feature averaging \citep{foster2020improving} has been applied successfully.

\paragraph{Function contrastive learning}
In their recent paper, \citet{gondal2021Function} considered function contrastive learning (FCLR) which uses a self-supervised objective to learn representations of functions.
FCLR fits naturally into the \gls{CRESP} framework as an \emph{untargeted} approach that uses mean-pooling in place of our attention aggregation.
Conceptually, then, FCLR does not take account of targeted tasks, nor does it propose a method for targeted representation learning.

\paragraph{Noise contrastive meta-learning}
\citet{ton2019noise} proposed an approach for conditional density estimation in meta-learning, motivated by multi-modal reconstruction.
Like targeted \gls{CRESP}, their method targets the unnormalized likelihood ratio \eqref{eq:targeted_lr}.
They use a noise contrastive \citep{gutmann2010noise} training objective with an explicitly defined `fake' distribution that is different from the \gls{CRESP} training objective.
Their primary method, MetaCDE, uses conditional mean embeddings to aggregate representations, unlike our attentive aggregation.
This means that, when using it as a baseline within our framework, MetaCDE does not form a fixed-dimensional representation of contexts, and so cannot be applied to untargeted tasks. 
They also proposed MetaNN, a purely neural version of their main approach.

\section{Experiments}
\label{sec:experiments}
We consider three different stochastic processes and downstream tasks which possess high-dimensional observations or complex noise distributions: 1) inferring parameters of periodic functions, 2) classifying 3D objects and 3) predicting collisions in a dynamical process.
We compare several models summarized in \cref{tab:methods} to learn representations of these stochastic processes.
All models share the same core encoder architecture. %
Please refer to \cref{sec:exp_details} for full experimental details.
\begin{table}[h!]
	\vspace{-1.2em}
	\centering
	\caption{
		Comparison of models used in at least one experiment in \cref{sec:experiments}.
		\label{tab:methods}}
		\vspace{.1em}
		\resizebox{1.\textwidth}{!}{
		\begin{tabular}{lcccccc}
		\toprule
		Criteria & \acrshort{CNP} & \acrshort{ACNP} & \acrshort{FCLR} & MetaCDE  & Targeted \gls{CRESP} & Untargeted \gls{CRESP}   \\
		\cmidrule(r){1-1}\cmidrule(lr){2-7}
		{\bf Targeted} & No & No & No & Yes & Yes & No   \\
		{\bf Reconstruction} & Yes & Yes & No & No & No & No   \\
		{\bf Attention} & No & Yes & No & No & Yes & Yes   \\
		\bottomrule
		\end{tabular}
		}
		\vspace{-1.2em}
	\end{table}
\subsection{Sinusoids}
\begin{figure}[b]
	\centering
	\begin{subfigure}[b]{0.30\linewidth}
		\centering\includegraphics[width=\textwidth]{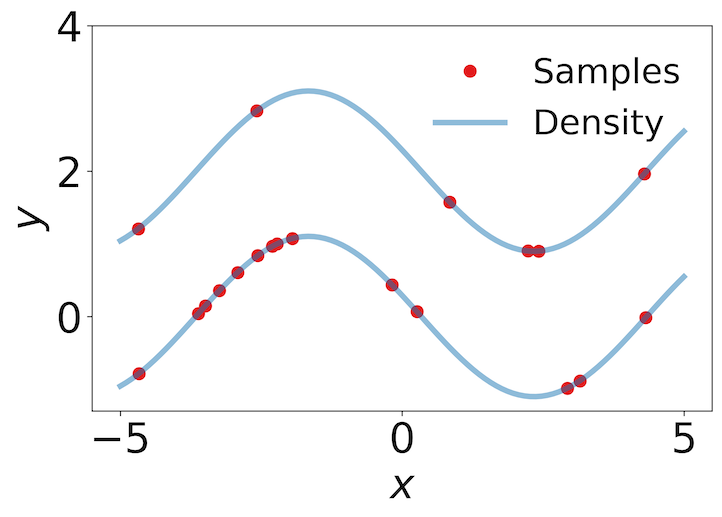}
		\caption{Stochastic process sample}
		\label{fig:sine_bimodal_noise}
	\end{subfigure}
	\hfill
	\begin{subfigure}[b]{0.34\linewidth}
		\centering\includegraphics[width=\textwidth]{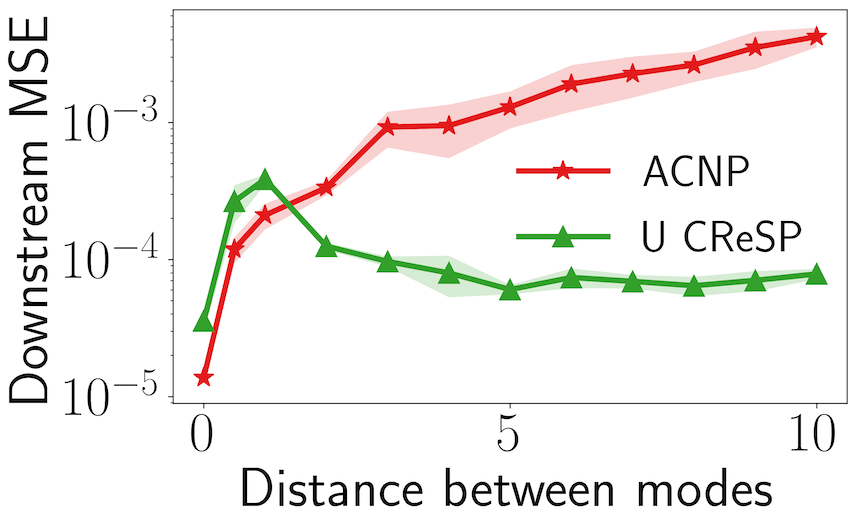}
		\caption{\gls{CRESP} vs \acrshort{ACNP}}
		\label{fig:sine_bimodal_noise_res}
	\end{subfigure}
	\hfill
	\begin{subfigure}[b]{0.34\linewidth}
		\centering\includegraphics[width=\textwidth]{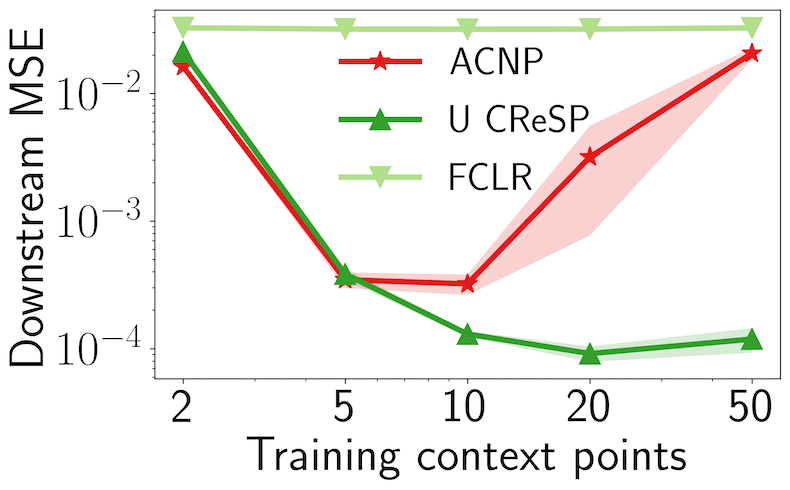}
		\caption{Effect of self-attention}
		\label{fig:sine_w_self_attn}
	\end{subfigure}
	\caption{
		We use \gls{CRESP} along with \acrshort{ACNP} and \gls{FCLR} to recover sinusoid parameters with a bi-modal likelihood.
		In each setting, we used $20$ test views to form representations of the entire training set and fitted a linear classifier to predict the function parameters.
		Encoders and decoder are MLPs.
		(a) Visualization of conditional likelihood $p(x|F,\cov)$. %
		(b)(c)
		Shaded areas represent $95\%$ confidence interval calculated using $6$ separately trained networks.
		We use the shorthand U = untargeted.
		In (b) we used $10$ training views and in (c) the distance between the modes is set to $2$.
		} 
		\label{fig:sine_noise}
\end{figure}

We first aim to demonstrate that reconstruction-based methods like \acrshort{CNPs} cannot cope well with a bi-modal noise process since their Gaussian likelihood assumption renders them misspecified.
We focus on a synthetic dataset of sinusoidal functions with both the observations and the covariates living in $\R$, i.e.\ $\Cov=\R$ and $\Obs=\R$.
We sample one dimensional functions $F \sim p(F)$ such that $F(x) = \alpha  \sin(2 \pi / T \cdot x  + \varphi)$ with random amplitude $\alpha \sim \mathcal{U}([0.5, 2.0])$, phase $\varphi \sim \mathcal{U}([0, \pi])$ and period $T=8$.
We break the uni-modality by assuming a bi-modal likelihood: $p(y|F,x) = 0.5~\delta_{F(x)}(y) + 0.5~\delta_{F(x) + \sigma}(y)$ (see \cref{fig:sine_bimodal_noise}).
Context points $x \in \Cov$ are uniformly sampled in $[-5, 5]$.
We consider the \emph{untargeted} downstream task of recovering the functions parameters $\lab=\{\alpha, \varphi\}$, %
and consequently put to the test our \emph{untargeted} \gls{CRESP} model along with \gls{FCLR} and ACNP.
We train all models for $200$ epochs, varying the distance between modes and the number of training context points.
We observe from \cref{fig:sine_bimodal_noise_res} that for high intermodal distance, the ACNP is unable to accurately recover the true parameters as opposed to \gls{CRESP}, which is more robust to this bi-modal noise even for distant modes.
Additionally, we see in \cref{fig:sine_w_self_attn} that self-attention is crucial to accurately recover the sinusoids parameters, as the MSE is several order of magnitude lower for \gls{CRESP} than for \gls{FCLR}.
We also see that \gls{CRESP} is able to utilize a larger context better than ACNP. %

\subsection{ShapeNet}
\todo{Motivation for real world problem? occulus and co infers the $SE(3)$ element etc.}
\begin{wrapfigure}{r}{0.46\textwidth}
	\vspace{-1.2em}
	\centering
	\begin{subfigure}[b]{0.75\linewidth}
		\centering\includegraphics[trim=0 200 0 0,clip,width=1\textwidth]{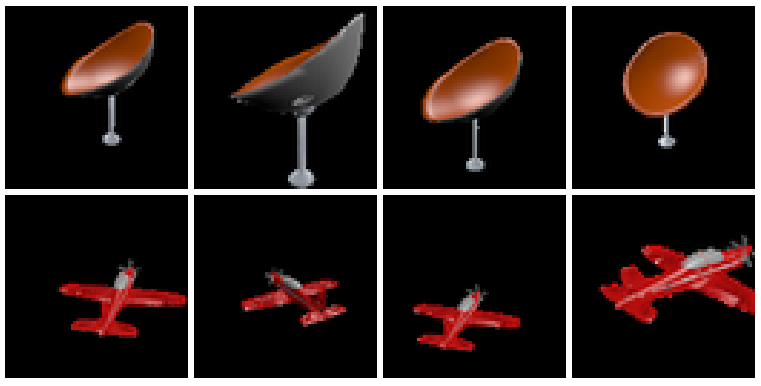}
	\end{subfigure}
	\begin{subfigure}[b]{0.75\linewidth}
		\centering\includegraphics[trim=0 0 1 0,clip,width=1\textwidth]{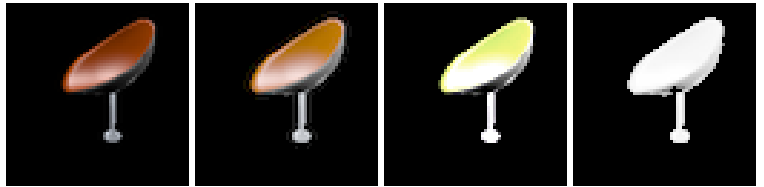}
	\end{subfigure}
	\caption{
		The ShapeNet dataset can be seen as a stochastic process: the covariate $\cov$ is the viewpoint and the observation $\obs$ is an image of the object from that viewpoint.
		[Top] We illustrate an object viewed from 4 random viewpoints.
		[Bottom] We show varying strengths of colour distortion applied to the same observation, the lefthand column is no distortion.
	}
	\label{fig:shapenet}
	\vspace{-1.5em}
\end{wrapfigure}

We apply \gls{CRESP} to ShapeNet  \citep{chang2015ShapeNet}, a standard dataset in the field of 3D object representations.
Each 3D object can be seen as a realization of a stochastic process with covariates $\cov$ representing viewpoints.
We sample random viewpoints involving both orientation and proximity to the object, with observations $\obs$ being $64\times 64$ images taken from that point.
We also apply randomized colour distortion as a noise process on the 2D images (see \cref{fig:shapenet}). As the likelihood of this noise process is not known in closed from, this should present a particular challenge to explicit likelihood driven models.
The downstream task for ShapeNet is a 13-way object classification which associates a single label with each realization---an untargeted task.
\begin{figure}
	\vspace{-0.5em}
	\centering
	\begin{subfigure}[b]{0.32\linewidth}
		\centering\includegraphics[width=1\textwidth]{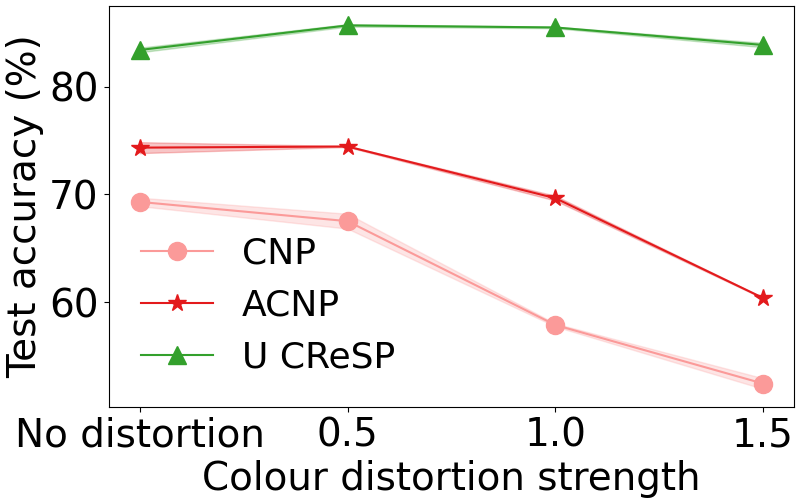}
		\caption{\gls{CRESP} vs reconstructive}
		\label{fig:shapenet_cresp_cnp}
	\end{subfigure}~
	\begin{subfigure}[b]{0.32\linewidth}
		\centering\includegraphics[width=\textwidth]{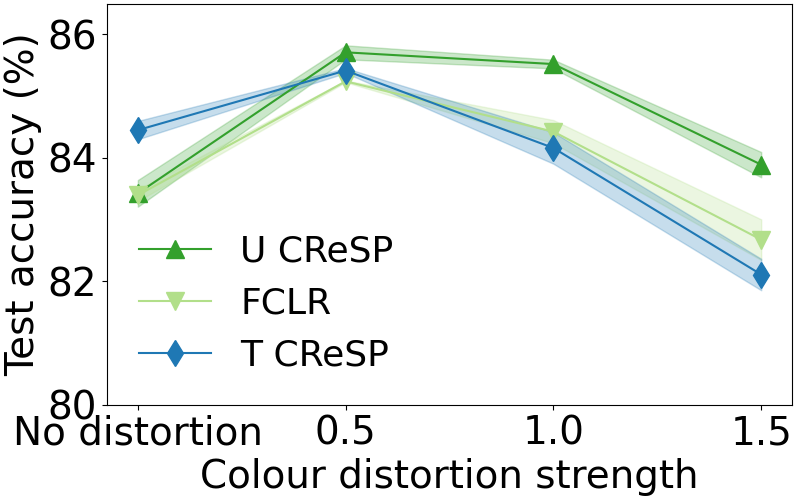}
		\caption{Contrastive methods}
		\label{fig:shapenet_contrastive}
	\end{subfigure}~
	\begin{subfigure}[b]{0.32\linewidth}
		\centering\includegraphics[width=\textwidth]{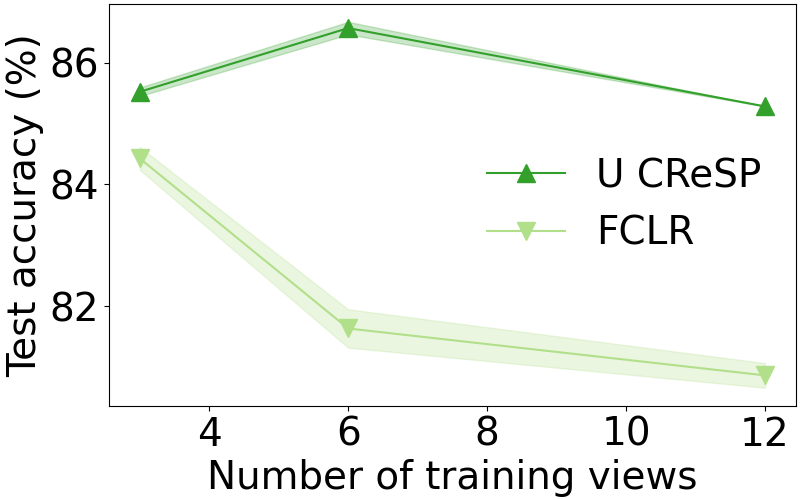}
		\caption{\gls{CRESP} vs FCLR}
		\label{fig:shapenet_train_views}
	\end{subfigure}
	\caption{We compare \gls{CRESP} with various baseline methods. In each case, we use 10 test views to form representations of the entire training set and fitted a linear classifier to predict ShapeNet object labels. Encoder networks were lightweight CNNs. In (a)(b) we used 3 training views, in (c) we used distortion strength 1. We present the test accuracy $\pm 1$ s.e. and we use the shorthand U = untargeted, T = targeted in figure legends. %
	}
	\label{fig:coldist}
	\vspace{-1.em}
\end{figure}

\paragraph{\gls{CRESP} outperforms reconstructive models}
Since the CNP learns by exact reconstruction of observations, we would expect it to struggle with high-dimensional image observations, and particularly suffer as we introduce colour distortion, which is a highly non-Gaussian noise process.
To verify this, we trained CNP and ACNP models, along with an attentive untargeted \gls{CRESP} model which we would expect to perform well on this task.
We used the same CNN observation processing network for each method, and an additional CNN decoder for the CNP and ACNP.
\cref{fig:shapenet_cresp_cnp} shows that \gls{CRESP} significantly outperforms both the CNP and ACNP, with reconstructive methods faring worse as the level of colour distortion is increased; \gls{CRESP} actually benefits from mild distortion.
\paragraph{\gls{CRESP} outperforms previous contrastive methods}
We next compare different contrastive approachs along two axes: targeted vs untargeted, and attentive vs pool aggregation. 
This allows a comparison with FCLR \citep{gondal2021Function}, which is an untargeted pool-based method.
\cref{fig:shapenet_contrastive} shows that no contrastive approach performs as badly as the reconstructive methods.
Untargeted \gls{CRESP} performs best, while the targeted method does less well on this untargeted downstream task. 
With our CNN encoders and a matched architecture for a fair comparison, FCLR does about as well as attentive targeted \gls{CRESP} and worse than the untargeted counterpart.
To further examine the benefits of the attention mechanism used in \gls{CRESP}, we vary the number of views used during training, focusing on untargeted methods. \cref{fig:shapenet_train_views} shows that as we increase the number of training views, the attentive method outperforms the non-attentive FCLR by an increasing margin. This indicates that careful aggregation and weighting of different views of each object is essential for learning the best representations.
The degradation in the performance of FCLR as more training views are used is likely due to a weaker training signal for the encoder as the self-supervised task becomes easier, this phenomenon also explains why \gls{CRESP} slightly \emph{decreases} in performance from 6 to 12 training views.

\begin{figure}[h!]
		\centering
		\begin{subfigure}[b]{0.34\linewidth}
			\centering\includegraphics[width=1\textwidth]{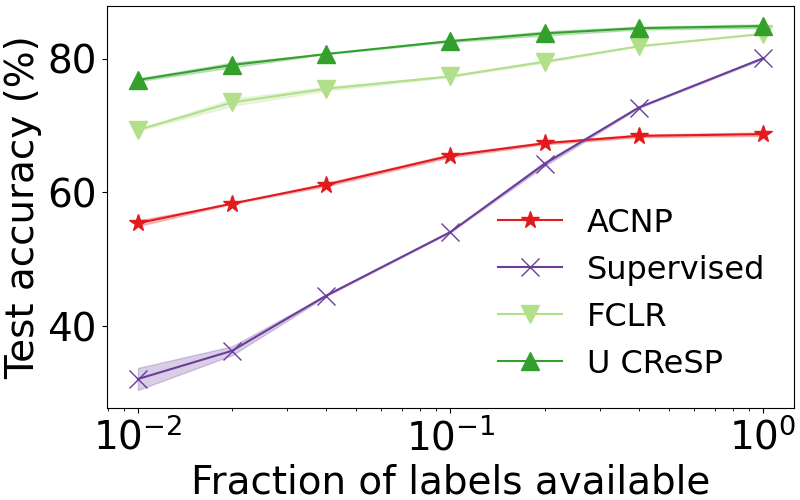}
			\caption{Semi-supervised evaluation}
			\label{fig:shapenet_semisupervision}
		\end{subfigure}~
		\hspace{2em}
		\begin{subfigure}[b]{0.34\linewidth}
			\centering\includegraphics[width=1\textwidth]{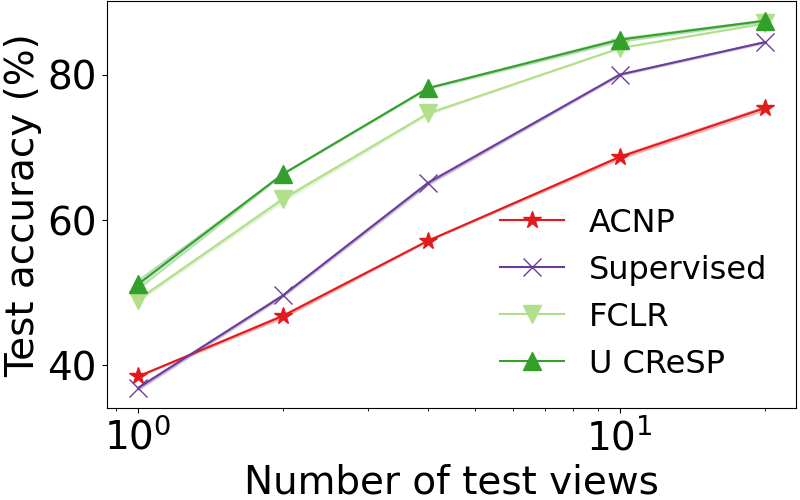}
			\caption{Test views}
			\label{fig:shapenet_test_views}
		\end{subfigure}~
		\caption{\gls{CRESP} for semi-supervised learning. We re-trained the final linear classifiers with different quantities of labelled data and number of test views, supervised learning trained the entire encoder architecture on the same labelled datasets. (a) We used 10 test views, (b) We used 100\% of labels. Other settings were as in \cref{fig:coldist}.
		}
		\label{fig:semisupervised}
	\end{figure}

\paragraph{\gls{CRESP} benefits from improved label efficiency}
We compare \gls{CRESP} with semi-supervised learning that does not use any pre-training, but instead trains the entire architecture on the labelled dataset. In \cref{fig:shapenet_semisupervision} we see that pre-training with \gls{CRESP} can outperform supervised learning on the same fixed dataset at \emph{every label fraction including 100\%}. Another axis of variation in the stochastic process setting is the number $C$ of views aggregated at test time. In \cref{fig:shapenet_test_views}, we see that performance increases across the board as we make more views available to form test representations, but that \gls{CRESP} performs best in all cases.
\subsection{Snooker dynamical process over images}
\begin{wrapfigure}{r}{0.40\textwidth}
	\vspace{-1.0em}
	\begin{subfigure}[b]{\linewidth}
		\centering
		\includegraphics[trim=-25 0 0 0,clip,width=1\linewidth]{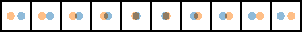} 
		\caption{2D images associated with target times.}
		\label{fig:snooker_target}
	\end{subfigure}
	\begin{subfigure}[b]{\linewidth}
		\includegraphics[trim=0 6 0 0,clip,width=1\linewidth]{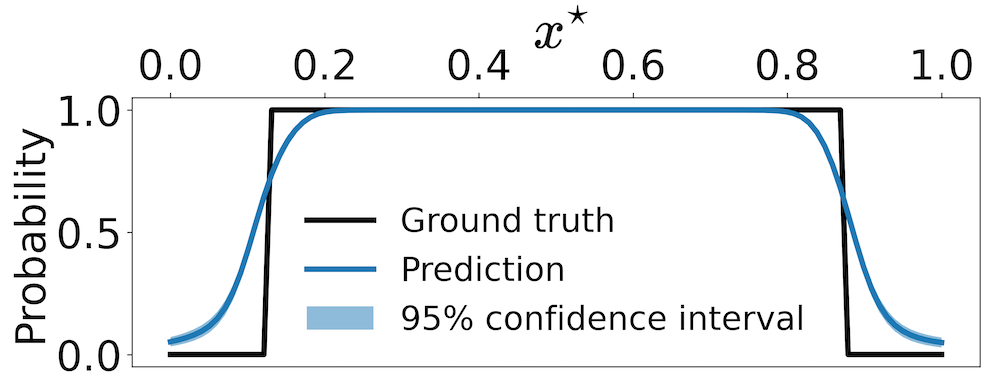}
		\caption{Probability of overlap.}
		\label{fig:snooker_prob}
	\end{subfigure}
	\caption{
		We assess the capacity of targeted \acrshort{CRESP} to smoothly predict whether the objects are overlapping at a given time $\cov^\star$ given a context set of size $5$.
		[Top] 2D images associated to $t\in[0,1]$.
		[Bottom]
		Confidence interval is computed over $50$ random contexts and $6$ trained models.
	}
	\label{fig:snooker}
	\vspace{-1.25em}
\end{wrapfigure}
We now focus on the setting where downstream tasks \emph{depend on the covariate $\covnew$}, i.e.\ \emph{targeted} downstream tasks. %
In particular, we consider a dynamical system that renders 2D images of two objects with constant velocities and evolving through time as illustrated in \cref{fig:snooker_target}.
The objects are constrained in a $1 \times 1$ box and collisions are assumed to result in a perfect reflection.
The observation space $\Obs$ is consequently the space of $28 \times 28$ RGB images, whilst the covariate space is $\R$, representing time.
We consider the downstream task of predicting whether the two objects are overlapping at a given time $\cov^\star=t$ or not. %
This experiment aims to reproduce, in a stripped-down manner, the real world problem of collision detection.
Even though the object's position can be expressed in closed-form, it is non trivial to predict the 2D image at a specific time given a collection of snapshots.
We expect \emph{targeted} \gls{CRESP} to be particularly well-suited for such a task since the model is learning to form and match a targeted representation to the representation of the ground truth observation thorough the \emph{unsupervised} task.
\paragraph{\gls{CRESP} outperforms reconstructive and previous contrastive methods}
Alongside targeted \gls{CRESP}, we consider the \acrshort{CNP}, \acrshort{FCLR} and MetaCDE models.
They are trained for $200$ epochs, 
with contexts of $5$ randomly sampled pairs $\left\{\obs_i = F(\cov_i), \cov_i \sim \mathcal{U}([0, 1]) \right\}$.
The encoder is a ResNet18 \citep{he2016deep}.
We found that self-attention did not seem to help any method for this task, so we report un-attentive models. 
Both \gls{CNP} and \acrshort{FCLR} learn \emph{untargeted} representations 
during the unsupervised task. We thus feed the downstream linear classifier with the concatenation $\{\bc,\cov^\star\}$.
Conversely, targeted \gls{CRESP} and MetaCDE directly produce a \emph{targeted} representation $\hat{\bc} = h(\cov^\star, \bc)$ (see \cref{fig:targeted_diag}). The downstream classifier can simply rely on $\hat{\bc}$ to predict the overlap label $\lab^\star$.
We consequently expect such a targeted representation $\hat{\bc}$ to be better correlated with the downstream label than untargeted representations $\bc$.

We observe from \cref{tab:snooker} that targeted \gls{CRESP} significantly outperforms both likelihood-based and previous contrastive methods, though MetaCDE outperforms both untargeted methods (\gls{CNP} and \acrshort{FCLR}).
This highlights the need for the \emph{targeted} contrastive loss from \cref{eq:targeted_cresp_loss} along with a flexible \emph{target head} $h$ to learn targeted representations. %
Additionally, we observe that in the absence of a noise process, \gls{CNP} performs as well as \acrshort{FCLR}.
We further investigate the quality of the learned targeted representations.
To do so, given a fixed context we make an overlap prediction at different points in time as shown in \cref{fig:snooker_prob}.
We observe that targeted \gls{CRESP} has successfully learned to smoothly predict the overlap label, but also to be uncertain when the overlap is ambiguous.
Thus \gls{CRESP} can successfully interpolate and extrapolate the semantic feature of interest (overlap) without reconstruction.

\begin{table}[h!]
	\centering
	\caption{
		We examine how well learned representations can predict whether the two snooker balls overlap at randomly sampled test times.
		$95\%$ confidence intervals were computed over $6$ runs. %
		\label{tab:snooker}}
		\vspace{.1em}
		\begin{tabular}{lcccc}
		\toprule
		& \gls{CNP} & \acrshort{FCLR} & Targeted \gls{CRESP} & MetaCDE \\ \midrule
		{\bf Accuracy (\%)} & ${85.3}_{\pm 0.5}$ & ${85.6}_{\pm 0.3}$ & $\bm{96.8}_{\pm 0.1}$ & ${87.7}_{\pm 0.3}$ \\
		\bottomrule
		\end{tabular}
\end{table}

\section{Discussion} \label{sec:discussion}
\paragraph{Limitations}
Our method directly learns representations from stochastic processes, without performing reconstruction on the observations, thus if one requires prediction in the observation space $\Obs$ then our method cannot be directly applied.
Whilst our method is tailor made for a setting of limited \emph{labelled} data, we require access to a large quantity of \emph{unlabelled} data to train our encoder network.

In this work, 
we do not place uncertainty over context representations. %
Learning stochastic embeddings would have the primary benefit of producing correlated predictions at two or more covariates, similarly to \glspl{NP}.
As there is no trivial nor unique way to extend the InfoNCE loss to deal with distributions \citep[e.g.][]{wu2020simple}, we leave such an extension of our method to future work.

\paragraph{Future applications}
One potential use of \gls{CRESP} is to generate representations that can be used for reinforcement learning, following the approach of \citet{eslami2018Neural}.
One of the key differences between real environments and toy environments is the presence of high-dimensional observations with naturalistic noise. This is a case where the contrastive approach can bring an edge because naturalistic noise significantly damages explicit likelihood methods, but \gls{CRESP} continues to perform well with more distortion.

\paragraph{Conclusion}
In this work, we introduced a framework for learning contrastive representation of stochastic processes (\gls{CRESP}).
We proposed two variants of our method specifically designed to effectively tackle \emph{targeted} and \emph{untargeted} downstream tasks.
By doing away with exact reconstruction, \gls{CRESP} directly works in the representation space, bypassing any challenge due to high dimensional and multimodal data reconstruction.
We empirically demonstrated that our methods are effective for 
dealing with multi-modal and naturalistic noise processes, 
and outperform previous contrastive methods for this domain on a range of downstream tasks.

\section*{Acknowledgments}
We would like to thank Yann Dubois and Jef Ton for valuable discussions.
We also thank Hyunjik Kim, Neil Band and Lewis Smith for providing feedback on earlier versions of the paper.
EM research leading to these results received funding from the European Research Council under the European Union’s Seventh Framework Programme (FP7/2007- 2013) ERC grant agreement no. 617071 and he acknowledges Microsoft Research and EPSRC for funding EM’s studentship.
AF gratefully acknowledges funding from EPSRC grant no. EP/N509711/1.

\bibliography{references}

\newpage
\begin{appendices}

\section{Broader impact} \label{sec:broader_impact}
The work presented in this paper focuses on the learning of representations for stochastic processes.
Applications in the field of computer vision could lead to better understanding of 3D scenes. Such applications could in turns lead to improved safety in products such as self-driving cars, as well as improved performance in areas such as medical imaging. 
Nonetheless, as with any computer vision technique, it might also be used in a way that carries societal risk.
As a foundational method, our work inherits the broader ethical aspects and future societal consequences of machine learning in general.

\section{Additional background}

\paragraph{Neural Processes}
Neural processes (NPs) learn a neural approximation $q\left(\obsnew|\covnew, (\cov_i,\obs_i)_{i=1}^C\right)$ to the posterior predictive distributions for stochastic processes given in \cref{eq:true_posterior_predictive}.
To create an efficient neural network architecture, the NP family use the fact that the posterior predictive distribution is unchanged under a permutation of the order $1,...,C$ of the context points.
The CNP combines representations of the observed data $\cov_{1:C},\obs_{1:C}$ into a \emph{context representation} $\bc$. To respect the permutation-invariance property, the CNP representation is of the form 
$\bc = \sum_{c} g_\text{enc}(\cov_c,\obs_c)$
where $g_\text{enc}: \Cov\times\Obs \rightarrow \C$ is an encoder.
The CNP predictions are then given by 
\begin{equation}
q\left(\obsnew|\covnew, (\cov_i,\obs_i)_{i=1}^C\right) =  p_\theta(\obsnew|\bc,\covnew)
\end{equation}
where $p_\theta(\cdot|\bc,\cov)$ is an explicit likelihood, conventionally a Gaussian with mean and variance given by a neural network applied to $\bc,\cov$.
The CNP model is then trained by maximum likelihood, i.e.\ by minimizing the following conditional log probability
\begin{equation}
	\mathcal{L}^\text{CNP} = - \E_{F} \left[ \E_{\cov,\obs} \left[\log q(\obsnew|(\cov_i,\obs_i)^C_{i=1},\covnew) \right] \right].
\end{equation}
Recall that the NP, unlike the CNP, includes an additional random variable $\bu$.
We can in fact view $\bu$ as a finite dimensional approximation to $F$ in \eqref{eq:true_posterior_predictive}.
In NPs, the random variable $\bu$ is sampled from an approximate posterior $q\left(\bu|(\cov_i,\obs_i)_{i=1}^C\right)$.
The NP constructs the approximate posterior so that it is invariant to the order of the context, by using a sum pooling approach to aggregate the context.
In order to learn this distribution, the NP introduces a modified training objective.
Considering a context set $(\cov_i,\obs_i)_{i=1}^C$ and target set $(\cov^\star_i,\obs^\star_i)_{i=1}^T$, the NP training loss \citep{garnelo2018neural} is
\begin{equation}
	-\E\left[\E_{q\left(\bu|(\cov_i,\obs_i)_{i=1}^C, (\cov^\star_i,\obs^\star_i)_{i=1}^T \right)}\left[ \sum_{i=1}^T \log q \left( \obs^\star_i | \cov^\star_i, \bu \right) + \log \frac{q\left(\bu|(\cov_i,\obs_i)_{i=1}^C \right)}{q\left(\bu|(\cov_i,\obs_i)_{i=1}^C, (\cov^\star_i,\obs^\star_i)_{i=1}^T \right)}  \right]\right]
\end{equation}
where $q \left( \obs^\star | \cov^\star, \bu \right)$ is the explicit likelihood model, typically a Gaussian, as in the CNP.
The outer expectation is with respect to the data $F, \cov,\obs$.

\paragraph{Attentive Neural Processes}
The ANP~\citep{kim2019attentive} introduced attention into the NP family in two different ways: \emph{self-attention} applies to the context to create context-aware representations of each context pair $(\cov_i,\obs_i)$; \emph{cross-attention} allows the ANP to attend to different components of the context depending on the target covariate $\cov^\star$.
These result in a representation $\hat{\bc}(\cov_{1:C}, \obs_{1:C}, \cov_i^\star)$ that depends on $\cov^\star$.
As with the NP, the ANP can include a latent variable $\bu$ to be sampled under a distribution that depends on $(\cov_i,\obs_i)_{i=1}^C$, self-attention can be used to generate the approximate posterior for $\bu$ in this case.
The overall training loss for the ANP is
\begin{equation}
\begin{split}
\mathcal{L}^\text{ANP} = -\E_{F, \cov,\obs}\bigg[ \E_{q\left(\bu|(\cov_i,\obs_i)_{i=1}^C \right)}\left[ \sum_{i=1}^T \log q \left( \obs^\star_i | \cov^\star_i, \bu, \hat{\bc}(\cov_{1:C}, \obs_{1:C}, \cov_i^\star) \right)\right] \\
\quad - \text{KL}\left[q\left(\bu | (\cov^\star_i,\obs^\star_i)_{i=1}^T \right) \| q\left(\bu|(\cov_i,\obs_i)_{i=1}^C \right)  \right]\bigg]
\end{split}
\end{equation}
where $q \left( \obs^\star_i | \cov^\star_i, \bu, \hat{\bc}(\cov_{1:C}, \obs_{1:C}, \cov_i^\star)\right)$ is the explicit likelihood model in this case.
We refer to the ANP model without the latent $\bu$ as the ACNP, for which the training loss is simply
\begin{equation}
\mathcal{L}^\text{ACNP} = -\E_{F, \cov,\obs}\left[
\sum_{i=1}^T \log q \left( \obs^\star_i | \cov^\star_i, \hat{\bc}(\cov_{1:C}, \obs_{1:C}, \cov_i^\star) \right)\right].
\end{equation}

\paragraph{Transformer attention}
The Image Transformer \citep{parmar2018image} used an attention mechansim based on multi-head self-attention \citep{vaswani2017attention}.
To describe this attention using our notation, suppose that $\br_1,\dots,\br_C$ are intermediate representations of pairs $(\cov_1,\obs_1),\dots,(\cov_C,\obs_C)$.
Then the $i$th representation $\br_i'$ in the next layer of representations is computed as follows.
We apply a query linear operator $W_q$ to $\br_i$ and a key linear operator $W_k$ to $\br_j$ for $j=1,\dots,C$.
We form a normalized set of weights
\begin{equation}
	w_{ij} = \frac{\exp\left(W_q\br_i \cdot W_k \br_j / \sqrt{d}\right)}{\sum_j \exp\left(W_q\br_i \cdot W_k \br_j  / \sqrt{d}\right)}
\end{equation}
where $d$ is the dimension of $\br_i$.
We then form a value as a weighted sum of existing representations, transformed with a value linear operator $W_v$ to give 
\begin{equation}
	\tilde{\br}_i = \sum_j w_{ij}W_v\br_j.
\end{equation}
To convert $\tilde{\br}_i$ to $\br_i'$, we apply dropout, a residual connection (i.e.~we add the original $\br_i$) and layer normalization \citep{ba2016layer}. Then we apply a second fully connected layer with residual connection and layer norm to give $\br_i'$.

\section{Method details}

\subsection{Downstream Tasks for Stochastic Processes} \label{sec:app:downstream}

We provide some additional details on targeted and untargeted tasks.
For a targeted task, we extend the stochastic process of Section~\ref{sec:sps} by introducing a second conditional distribution $p(\lab|F,\cov)$. We assume that the joint distribution over observations $\obs_{1:C}$ and labels $\lab_{1:C}$ is given by
\begin{equation}
\label{eq:targeted}
p\left(\obs_{1:C}, \lab_{1:C}|\cov_{1:C}\right) = \int p(F)\prod_{i=1}^C p(\obs_i|F, \cov_i)p(\lab_i|F,\cov) \ dF,
\end{equation}
implying that the predictive density of the label $\lab^\star$ at $\covnew$ given the context $\{(\cov_i,\obs_i)_{i=1}^C\}$ is
\begin{equation}
\label{eq:targeted_prediction}
p\left(\lab^\star|\covnew, (\cov_i,\obs_i)_{i=1}^C\right) = \frac{\int p(F)p(\lab^\star|F,\covnew)\prod_{i=1}^C p(\obs_i|F, \cov_i) \ dF}{\int p(F)p(\lab^\star|F,\covnew) \ dF }.
\end{equation}
In \gls{CRESP}, we estimate this by forming a targeted representation $\hat{\bc}$ of $(\cov_i,\obs_i)_{i=1}^C$ and $\covnew$, and fitting a linear model $q(\lab|\hat{\bc})$.

For untargeted tasks, there is one $\lab$ sampled along with the entire realization $F$ via a conditional distribution $p(\lab|F)$, giving the joint distribution
\begin{equation}
\label{eq:app:untargeted}
p(\obs_{1:C}, \lab|\cov_{1:C}) = \int p(F)p(\lab|F)\prod_{i=1}^C p(\obs_i|F, \cov_i) \ dF.
\end{equation}
This means that we can predict $\lab$ using the context $\{(\cov_i,\obs_i)_{i=1}^C\}$ using the predictive distribution
\begin{equation}
\label{eq:untargeted_prediction}
p\left(\lab^\star|(\cov_i,\obs_i)_{i=1}^C\right) = \frac{\int p(F)p(\lab^\star|F)\prod_{i=1}^C p(\obs_i|F, \cov_i) \ dF}{\int p(F)p(\lab^\star|F) \ dF }.
\end{equation}
In \gls{CRESP}, we estimate this using a representation $\bc$ of $(\cov_i,\obs_i)_{i=1}^C$; we fit a linear model $q(\lab|\bc)$.

\section{Experimental details} \label{sec:exp_details}

We provide below all necessary details to understand and reproduce the empirical results obtain in \cref{sec:experiments}.
Hyperparameters are summarized in \cref{tab:hyperparameters}.
Models were implemented in PyTorch~\citep{paszke2017automatic}.
For downstream tasks we fit linear models with L-BFGS~\citep{liu1989On}, we applied L2 regularization to the weights.
Our code is available at \url{github.com/ae-foster/cresp}. 

\begin{table}[h]
	\centering
	\caption{Hyperparameters used for the different experiments.}
	\label{tab:hyperparameters}
	\begin{tabular}{lrrr}
		\toprule
		Parameter           & Sinusoids        & ShapeNet         & Snooker       \\
		\cmidrule(r){1-1}\cmidrule(lr){2-4}
		Covariate space $\Cov$  & $\R$             & $\R^{15}$        & $\R$    \\
		Observation space $\Obs$ & $\R$             & RBG 64x64 images & RBG 28x28 images  \\
		Dataset sizes       & 17.6k/2.2k/2.2k  & 26270/8756/8756     & 15k/3k/20k  \\
		Observation Net     & Id               & CNN              & ResNet18   \\
		Covariate Net       & Id               & Id               & Id   \\
		Encoder Net         & MLP              & Gated            & Gated   \\
		Decoder model       & MLP              & CNN              & DCGAN   \\
		Attention           & 2 transformer layers  & 2 transformer layers  &  \\
		Target network      &                  & Gated            & MLP   \\
		Training views      & 10               & 3                & 5  \\
		Test views          & 20               & 10               & 9  \\
		Representation dim  & 512              & 512              & 512  \\
		Projection dim      & 128              & 128              & 128  \\
		Training batch size & 256              & 512              & 256  \\
		Training epochs     & 200              & 10               & 200   \\
		Optimizer           & Adam             & LARS             & Adam    \\
		Scheduler           & Cosine           & Cosine + Ramp    & Cosine + Ramp     \\
		Scheduler Ramp length &                & 10               & 10     \\
		Learning rate       & 3e-4             & 2e-1             & 2e-3 \\
		Momentum            & 0.9              & 0.9              & 0.9 \\
		Weight decay        & 1e-6             & 1e-6             & 1e-6 \\
		Temperature $\tau$  & 0.5              & 0.5              & 0.5 \\
		Downstream L2 regularization & 1e-6    & 1e-3             & 1e-3 \\
		\bottomrule
	\end{tabular}
\end{table}

\subsection{CO$_2$ emissions}
Experiments were conducted using a private infrastructure, which has an estimated carbon efficiency of 0.188 kgCO$_2$eq/kWh \footnote{Average carbon intensity in March, April and June in the Great Britain. Source \url{https://electricityinfo.org/carbon-intensity-archive}.}.
An estimated cumulative 1000 hours of computation was performed on hardware of type RTX 2080 Ti (TDP of 250W), or similar such as RTX 1080 Ti.
Total emissions are estimated to be 47 kgCO$_2$eq. %
Estimations were conducted using the \href{https://mlco2.github.io/impact#compute}{Machine Learning Impact calculator} presented in \cite{lacoste2019quantifying}.

\subsection{Sinusoids dataset} \label{sec:sine_detail}
\paragraph{Data}
We sample unidimensional functions $F \sim p(F)$ such that $F(x) = \alpha  \sin(2 \pi / T \cdot x  + \varphi)$ with random amplitude $\alpha \sim \mathcal{U}([0.5, 2.0])$, phase $\varphi \sim \mathcal{U}([0, \pi])$ and period $T=8$.
We assume a bimodal likelihood: $p(y|F,x) = 0.5~\delta_{F(x)}(y) + 0.5~\delta_{F(x) + \sigma}(y)$.
Context points $x \in \Cov$ are uniformly sampled in $[-5, 5]$.

\paragraph{Architectures}
Since both the covariate and observation variables are unidimensional, we do not preprocess them, i.e.\ $g_\text{cov}=\text{Id}$ and $g_\text{obs}=\text{Id}$.
For the encoder--processing $g_\text{enc}(g_\text{cov}(\cov), g_\text{obs}(\obs))$--we rely on an \gls{MLP} with $3$ hidden layer of $512$ hidden units.
For reconstructive methods (CNP and ACNP), the decoder is also parametrized by an \gls{MLP} with $512$ hidden units and $3$ hidden layers.

\subsection{Shapenet dataset} \label{sec:shapenet_detail}
\paragraph{Data} We utilize the renderings of ShapeNet objects provided in 3D-R$^2$N$^2$ \citep{choy20163d}. 
These renderings are constructed from different orientations. We also apply a random crop to each image to simulate a random proximity to the object. Specifically, we choose a random area from $U(0.08, 1)$ and then a random crop of that area.
This process is summarized by the PyTorch snippet
\begin{verbatim}
bounding_box = list(transforms.RandomResizedCrop.get_params(
    img, (0.08, 1), (1., 1.)
))
img = transforms.functional.resized_crop(
    img, *bounding_box, 64, Image.LANCZOS
)
\end{verbatim}
This means that the covariate $\cov$ representing the view consists of the angles describing the orientation of the render, and the bounding box. We apply additional featurization to $\cov$ described in the next section.
We also apply random colour distortion of strength $s$ as a noise process on the images $\obs$. Inspired by the colour distortion of \citet{chen2020simple} we apply randomized brightness, contrast, saturation, hue and gamma adjustment (see our code for the exact implementation).

\paragraph{Feature processing} We process the covariate $\cov$ as follows. For the azimuthal angle $\theta$, we use $\sin(n\theta),\cos(n\theta)$  for $n=1,2,3$ and the original angle (7 features). We include the elevation and distance of the R$^2$N$^2$ render without additional features (2 features): in practice these vary little in this dataset.
We include the bounding box mid-point and area as additional features, along with the four corners of the bounding box (6 features). 
All told, this gives a covariate of dimension 15. We finally apply normalization to the covariate so that each component has mean 0 and variance 1 over the entire dataset.
To images $\obs$ we apply a linear rescaling that means each channel has mean 0 over the dataset.

\paragraph{Learning set-up and downstream tasks}
For unsupervized learning, we resample the view and distortion randomly each time an object is encountered.
For learning on downstream tasks, we fix a dataset of covariates, observations and labels, and learn exclusively from this fixed dataset without resampling views, providing a more realistic semi-supervised test case.
The labels are included in the dataset, but only utilized by our algorithm when we train downstream linear classifiers (except for the supervised baseline).
The following 13 categories are represented in our dataset: display (1095), watercraft (1939), bench (1816), telephone (1052), cabinet (1572), sofa (3173), rifle (2373), loudspeaker (1618), airplane (4045), table (8509), chair (6778), car (7496), lamp (2318).

\paragraph{Architectures}
For the observation network, we use a CNN described by the following PyTorch snippet
\begin{verbatim}
nn.Sequential(
    nn.Conv2d(num_channels, ngf // 8, 3, stride=2, padding=1, bias=False),
    nn.BatchNorm2d(ngf // 8),
    nn.LeakyReLU(),
    nn.Conv2d(ngf // 8, ngf // 4, 3, stride=2, padding=1, bias=False),
    nn.BatchNorm2d(ngf // 4),
    nn.LeakyReLU(),
    nn.Conv2d(ngf // 4, ngf // 2, 3, stride=4, padding=1, bias=False),
    nn.BatchNorm2d(ngf // 2),
    nn.LeakyReLU(),
    nn.Conv2d(ngf // 2, ngf, 3, stride=4, padding=1),
    nn.BatchNorm2d(ngf),
    nn.LeakyReLU(),
)
\end{verbatim}
and we set \texttt{ngf}$=512$.
For reconstructive methods (CNP and ACNP), we use a convolutional decoder of the following form
\begin{verbatim}
nn.Sequential(
    nn.UpsamplingNearest2d(scale_factor=2),
    nn.ConvTranspose2d(nz, ngf // 2, 2, stride=2, padding=0, bias=False),
    nn.BatchNorm2d(ngf // 2),
    nn.LeakyReLU(),
    nn.UpsamplingNearest2d(scale_factor=2),
    nn.ConvTranspose2d(ngf // 2, ngf // 4, 2, stride=2, padding=0, bias=False),
    nn.BatchNorm2d(ngf // 4),
    nn.LeakyReLU(),
    nn.ConvTranspose2d(ngf // 4, ngf // 8, 2, stride=2, padding=0, bias=False),
    nn.BatchNorm2d(ngf // 8),
    nn.LeakyReLU(),
    nn.ConvTranspose2d(ngf // 8, nc, 2, stride=2, padding=0),
)
\end{verbatim}
where \texttt{nz}$=512 + 15$, \texttt{ngf}$=512$, \texttt{nc}$=6$.
Finally, we extract three means and three standard deviations from the output at each pixel location for three colour channels, applying a sigmoid to the means (to put them in the correct range for image data) and a softplus transform to the standard deviations.

The gated unit that we use is as follows
\begin{verbatim}
class Gated(nn.Module):

    def __init__(self, in_dim, representation_dim):
        super(Gated, self).__init__()
        self.fc1 = nn.Linear(in_dim, representation_dim)
        self.fc2 = nn.Linear(in_dim, representation_dim)
        self.activation = nn.Sigmoid()

    def forward(self, x):
        representation = self.fc1(x)
        multiplicative = self.activation(self.fc2(x))
        return multiplicative * representation
\end{verbatim}
inspired by gated units that appear in \citet{hochreiter1997long,cho2014learning}.
The gated unit is utilized in two places: as the pair encoding (\cref{sec:representation}) that processes the covariate and observation features after concatenation, and as the target network for our targeted \gls{CRESP} implementation on ShapeNet.
We found that it slightly outperformed an MLP with a similar number of parameters.

\subsection{Snooker dataset} \label{sec:snooker_detail}
\paragraph{Data}
This synthetic dataset simulates a dynamical system with two objects evolving through time with constant velocities.
Formally, let's consider two objects at positions $\rvs_i$ at time $t$. A free object moving at velocity $\bv_i$ has position
	$\rvs_i(t) = \rvs_i(0) + \bv_i t$.
We now consider both objects constrained so that $0 \le \rvs_i \le 1$ and assume that collisions with the boundaries result in a perfect reflection.
The position of the particle can be expressed by the following formula
\begin{align} \label{eq:snooker_position}
	\tilde{s}_i(t) &= s_i(0) + v_it,\\
	\begin{split}
	s_i(t) &= (\lfloor \tilde{s}_i(t) \rfloor \mod 2) (1 - \tilde{s}_i(t) + \lfloor \tilde{s}_i(t) \rfloor) + (1 - \lfloor \tilde{s}_i(t) \rfloor \mod 2) (\tilde{s}_i(t) - \lfloor \tilde{s}_i(t) \rfloor)
	\end{split}
\end{align}
for $i=1,2$.

We then assume that we only have access to a 2D image $\obs$ of the state at time $\cov=t$ for a given realization $F$. %
We sample realizations $F \sim p(F)$ such that $\rvs_i(0) \sim \mathcal{U}([0,1]^2)$, and $\bv_i=v_0 \bm{\alpha}$ with $\bm{\alpha} \sim \mathcal{U}(\mathbb{S}^1)$ and $v_0 = 0.4$.
The objects are assumed to be non-interacting discs of radius $0.15$.

The downstream task is to predict whether the two objects are overloading at a given time, i.e.\ $\E_{p(\lab|F,\cov^\star=t)}[\lab]$ with $\lab=1$ if there is an overlap.
The objects position can be expressed at any time in closed-form (cf \cref{eq:snooker_position}), yet it is quite challenging to predict the 2D image at a specific time given a collection of snapshots.

\paragraph{Architectures}
For the observation network, we use a CNN described by the following PyTorch snippet
\begin{verbatim}
nn.Sequential(
    nn.Conv2d(nc, ngf, kernel_size=2, stride=2, bias=False),
    nn.BatchNorm2d(ngf),
    nn.ReLU(True),
    nn.Conv2d(ngf, 2 * ngf, kernel_size=2, stride=2, bias=False),
    nn.BatchNorm2d(2 * ngf),
    nn.ReLU(True),
    nn.Conv2d(2 * ngf, 4 * ngf, kernel_size=2, stride=2, bias=False),
    nn.BatchNorm2d(4 * ngf),
    nn.ReLU(True),
    nn.Conv2d(4 * ngf, nz, kernel_size=2, stride=2),
)
\end{verbatim}
where \texttt{ngf}$=64$ and \texttt{nc}$=3$.
For reconstructive methods (CNP and ACNP), we use a convolutional decoder inspired by DCGAN \citep{radford2016Unsupervised}, of the form
\begin{verbatim}
nn.Sequential(
    nn.ConvTranspose2d(nz, ngf * 4, 4, 1, 0, bias=False),
    nn.BatchNorm2d(ngf * 4),
    nn.ReLU(True),
    nn.ConvTranspose2d(ngf * 4, ngf * 2, 3, 2, 1, bias=False),
    nn.BatchNorm2d(ngf * 2),
    nn.ReLU(True),
    nn.ConvTranspose2d(ngf * 2, ngf, 4, 2, 1, bias=False),
    nn.BatchNorm2d(ngf),
    nn.ReLU(True),
    nn.ConvTranspose2d(ngf, 2 * nc, 4, 2, 1),
)
\end{verbatim}
where \texttt{nz}$=512 + 1$, \texttt{ngf}$=64$ and \texttt{nc}$=2*3$.
Similarly to \cref{sec:shapenet_detail}, we extract three means and three standard deviations from the output at each pixel location. %

For the encoder--processing $g_\text{enc}(g_\text{cov}(\cov), g_\text{obs}(\obs))$--we rely on the gated architecture described above in \cref{sec:shapenet_detail}.
For the target network $h$, which outputs the predictive representation $\hat{\bc} = h([\cov^\star, \bc])$, we rely on an MLP with $3$ hidden layers of $512$ units each.

\clearpage
\end{appendices}

\end{document}